\documentclass[pdflatex,sn-mathphys-num,iicol]{sn-jnl}

\usepackage{graphicx}
\usepackage{multirow}%
\usepackage{amsmath,amssymb,amsfonts}%
\usepackage{amsthm}%
\usepackage{mathrsfs}%
\usepackage[title]{appendix}%
\usepackage{xcolor}%
\usepackage{textcomp}%
\usepackage{manyfoot}%
\usepackage{booktabs}%
\usepackage{algorithm}%
\usepackage{algorithmicx}%
\usepackage{algpseudocode}%
\usepackage{listings}%
\usepackage{adjustbox}

\usepackage{array}
\usepackage{amsmath}
\usepackage{comment}
\usepackage{epstopdf}
\usepackage{subcaption}

\theoremstyle{thmstyleone}%
\theoremstyle{thmstyletwo}%

\theoremstyle{thmstylethree}%

\begin{document}

\title[Multimodal LLMs for Built Environment and Housing Attribute Assessment]{Leveraging Multimodal LLMs for Built Environment and Housing Attribute Assessment from Street-View Imagery}

\author*[1]{\fnm{Siyuan} \sur{Yao}}\email{syao2@nd.edu}

\author[1]{\fnm{Siavash} \sur{Ghorbany}}\email{sghorban@nd.edu}

\author*[1]{\fnm{Kuangshi} \sur{Ai}}\email{kai@nd.edu}

\author[1]{\fnm{Arnav} \sur{Cherukuthota}}\email{acheruku@nd.edu}

\author[1]{\fnm{Meghan} \sur{Forstchen}}\email{mforstch@nd.edu}

\author[1]{\fnm{Alexis} \sur{Korotasz}}\email{akorotas@nd.edu}

\author[1]{\fnm{Matthew} \sur{Sisk}}\email{msisk1@nd.edu}

\author[1]{\fnm{Ming} \sur{Hu}}\email{mhu1@nd.edu}

\author*[1]{\fnm{Chaoli} \sur{Wang}}\email{chaoli.wang@nd.edu}

\affil*[1]{\orgname{University of Notre Dame}, \orgaddress{\city{Notre Dame}, \state{IN}, \postcode{46556}, \country{USA}}}


\abstract{We present a novel framework for automatically evaluating building conditions nationwide in the United States by leveraging large language models (LLMs) and Google Street View (GSV) imagery. By fine-tuning Gemma 3 27B on a modest human-labeled dataset, our approach achieves strong alignment with human mean opinion scores (MOS), outperforming even individual raters on SRCC and PLCC relative to the MOS benchmark. To enhance efficiency, we apply knowledge distillation, transferring the capabilities of Gemma 3 27B to a smaller Gemma 3 4B model that achieves comparable performance with a 3× speedup. Further, we distill the knowledge into a CNN-based model (EfficientNetV2-M) and a transformer (SwinV2-B), delivering close performance while achieving a 30× speed gain. Furthermore, we investigate LLMs' capabilities for assessing an extensive list of built environment and housing attributes through a human–AI alignment study and develop a visualization dashboard that integrates LLM assessment outcomes for downstream analysis by homeowners. Our framework offers a flexible and efficient solution for large-scale building condition assessment, enabling high accuracy with minimal human labeling effort.}

\keywords{Built environment evaluation, housing attribute assessment, street-view imagery, computer vision, machine learning, multimodal large language models, visualization dashboard}



\maketitle

\section{Introduction}

The persistent shortage of affordable housing in the United States, aging infrastructure, and rising energy costs burden low-income households~\cite{Anacker-IJHP19}.
Retrofitting existing housing stock has emerged as a more feasible and cost-effective alternative to new construction, offering opportunities to improve thermal comfort, reduce utility bills, and mitigate health risks from extreme heat and cold~\cite{Hu-NSF25}.
Assessing the physical condition of buildings, especially the exterior envelope (i.e., façades), is critical to determining retrofit needs. However, audits remain resource-intensive, requiring manual inspections and data collection, limiting their practice at a nationwide scale. 

Advances in computer vision \cite{SG-ACMCS23,Ghorbany-BE24,Yao-ISVC24,Ghorbany-SR25} have enabled automated analysis of street-view imagery, such as \textit{Google Street View} (GSV) images, to identify \textit{passive design indicators} (PDIs) like window-to-wall ratios, shading devices, and exterior material types.
While these approaches leverage standard computer vision models (e.g., ResNet) or vision-language models (e.g., BLIP) to classify exterior features with high accuracy, their capacity to simultaneously consider diverse visual elements indicating various features (e.g., paint, window, and structural conditions) and provide interpretable assessments remains limited. \textit{Large language models} (LLMs), especially multimodal LLMs that integrate visual and textual reasoning, present a promising new frontier for interpretable and generalizable assessment. Aligning LLMs with human rater evaluations through fine-tuning on annotated street-view imagery could enable accurate and scalable assessments of building conditions, thereby supporting data-driven passive retrofit strategies across diverse urban contexts. 

This paper presents a novel method for fine-tuning multimodal LLMs to automate exterior condition assessments for residential buildings, leading to a pipeline that fine-tunes multimodal LLMs to interpret building exteriors, enabling scalable, automated, and cost-effective assessments. 
The extension of our prior ISVC 2025 work~\cite{Yao-ISVC25} includes the following. First, we expand the scope of assessment beyond a single housing condition to a more comprehensive set of residential context indicators, enabling richer and more holistic characterization of housing conditions. Second, we conduct a human–AI alignment study to evaluate and improve the consistency between multimodal LLM predictions and expert judgments, thereby enhancing reliability and interpretability. Finally, we develop an interactive visualization dashboard that integrates model outputs into an accessible interface, supporting downstream analysis and decision-making by homeowners and other stakeholders.

\section{Related Work}

Automated visual assessment of the built environment has become an increasingly active area of research, driven by the growing availability of street-view imagery and advances in machine learning. 
Traditional computer vision approaches, such as CNNs, have been employed to extract architectural features from images for estimating building condition~\cite{Hoang-CIN18,Amrouni-PCF24,Zou-ISPRS21}. 
However, these models often rely on narrow indicators, such as wall paint cracks, and struggle with low accuracy when integrating multiple factors due to limited generalizability.

More recently, vision-language models have demonstrated impressive zero-shot classification performance on visual tasks by leveraging cross-modal alignment between images and text.
These models have been applied to identify building components or construction materials~\cite{Yao-ISVC24}, as well as to estimate physical walkability~\cite{Liu-SIG23} and perceived safety~\cite{Wang-AGILE25}, directly from street-view imagery without needing extensive labeled data.
However, their effectiveness in structured assessment tasks, such as condition scoring guided by formal criteria, remains limited due to insufficient capacity for multi-factor reasoning.

There is a growing interest in using LLMs to extract features of the built environment from street-view images. 
Several recent studies~\cite{Cheng-VTC24,Li-arXiv24,Malekzade-CEUS25,Liang-arXiv25} have employed ChatGPT to capture detailed building information, including external features and nearby environmental elements. 
However, ChatGPT’s closed-source nature, lack of support for fine-tuning, and the high cost of API usage at scale limit its practicality for large datasets. 
Alternatively, fine-tuned open-source LLMs can be trained on labeled datasets for post-earthquake structural damage assessment, successfully performing tasks such as identifying damage severity and classifying affected components~\cite{Jiang-CACIE25}.

\begin{figure*}
\centering
\includegraphics[width=0.9\textwidth]{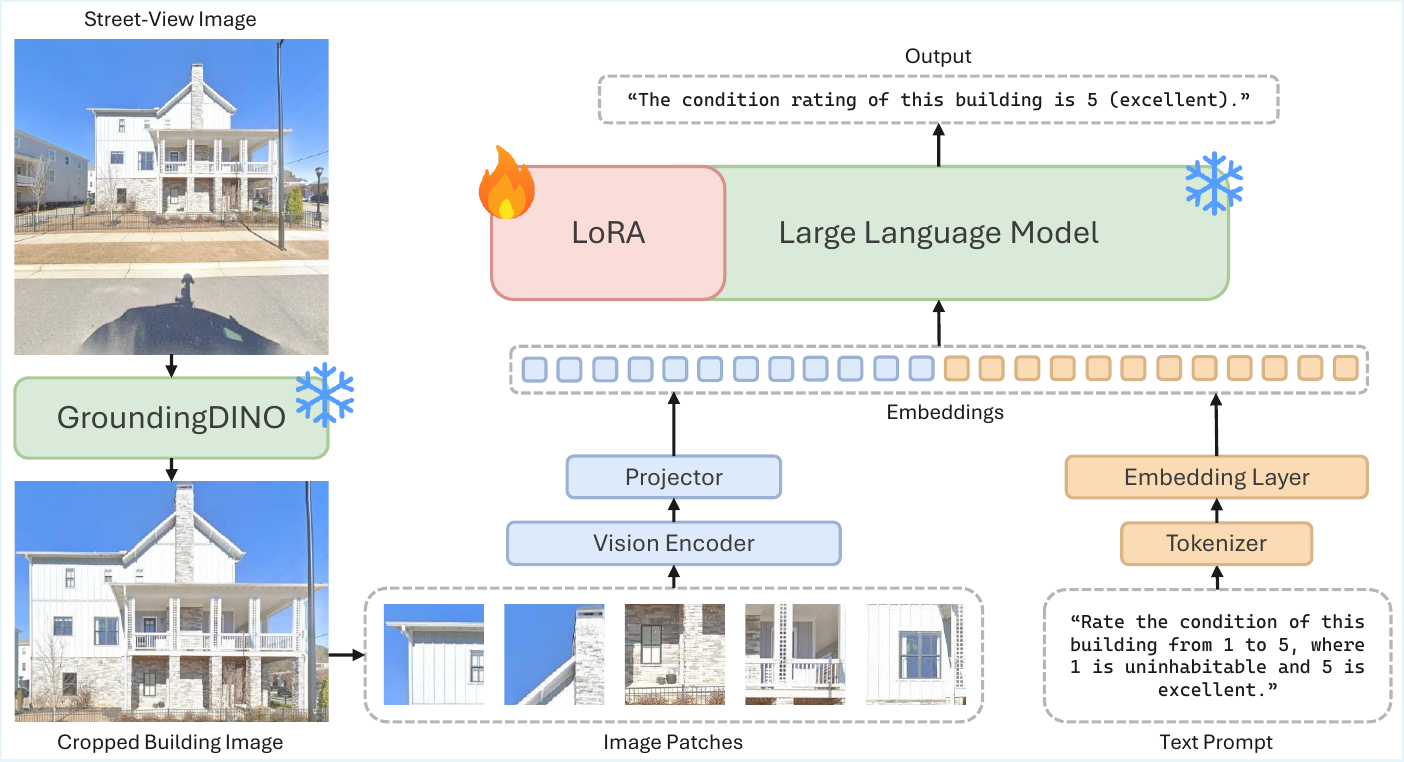}
\caption{Overview of our framework for evaluating building conditions.} 
\label{fig:overview}
\end{figure*}

Beyond feature extraction, recent work has explored LLMs as judges to approximate human judgment in subjective, multi-criteria tasks~\cite{gu2024survey,zheng2023llmjudge}, combining scalability with context-aware reasoning and achieving strong agreement with human preferences. However, concerns remain regarding reliability, bias, and prompt sensitivity, highlighting the need for careful validation and human alignment~\cite{shankar2024who,chehbouni2025neither}. Recent human-in-the-loop and active learning approaches have further emphasized the importance of incorporating human feedback to improve evaluation quality~\cite{gebreegziabher-etal-2025-leveraging}. Motivated by these findings, we adopt LLMs as judges for large-scale housing attribute assessment and conduct a human-LLM alignment study to ensure consistency with expert criteria.

In this work, we fine-tune open-source LLMs for building condition assessment by leveraging diverse visual features that reflect natural aging and property upkeep.
These include indicators like the state of windows, façade paint, and roof materials.
In addition, we leverage advanced closed-source multimodal LLMs (e.g., GPT-5.4, Gemini-3.1-Pro, and Claude-Opus-4.6) to extract a broader set of housing attributes beyond overall condition, and align their predictions with human expert annotations through a human–LLM agreement study.
Our system produces assessments consistent with expert judgment by aligning model outputs with \textit{mean opinion scores} (MOS) derived from multiple human raters.

\begin{figure*}
\centering
\includegraphics[width=\textwidth]{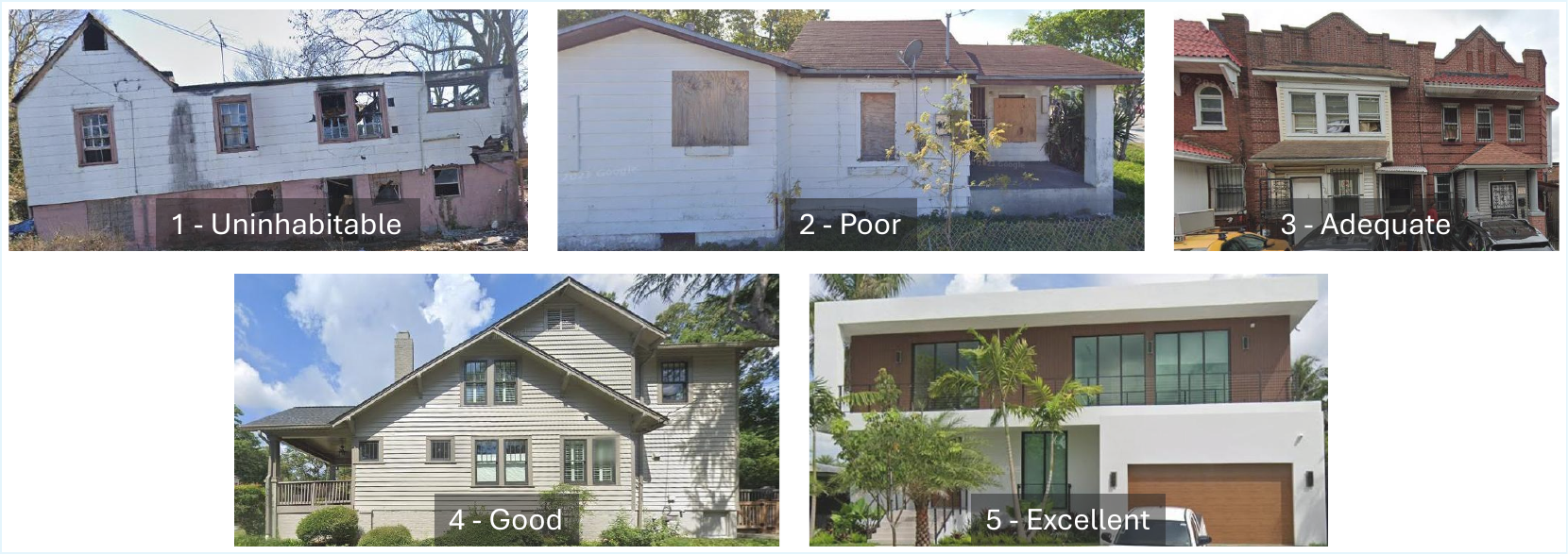}
\caption{Examples of buildings corresponding to each condition rating.} 
\label{fig:rating-sample}
\end{figure*}

\section{Our Approach}
\label{sec:approach}

We aim to evaluate the condition of a building from its GSV image. Experts in architectural research have established formal criteria for condition assessment using a five-point scale (higher ratings indicate better condition). Each rating is defined by detailed descriptors, covering various aspects such as the building's main structure, wall integrity, paint condition, roof, and window quality.

One possible approach is replicating human evaluation by training individual models for each component (e.g., roof, windows, façades), followed by a secondary algorithm to combine these assessments into an overall rating. However, this approach presents several limitations. First, many components may be partially occluded or absent from the image due to angle or visibility. Second, creating labeled datasets for each component would require significant manual effort from experts. Finally, integrating multiple separate models into a single scoring pipeline is complex and inefficient.

To address these challenges, we adopt a more direct strategy: evaluating the entire building holistically using a single model. This method assumes the model can effectively incorporate the full evaluation rubric and infer a corresponding rating. 
Following the method outlined in \cite{Yao-ISVC24}, and as illustrated in Figure~\ref{fig:overview}, we preprocess images using GroundingDINO to isolate and crop individual buildings from street-view scenes. As found in the experiment in Section~\ref{subsec:llm}, we choose Gemma 3~\cite{Gemma-arXiv24} as the multimodal LLM backbone to process each cropped image alongside a structured text prompt. In practice, in addition to the query, we include the formal rating criteria in the text prompt as follows:

\begin{itemize}
\item \textbf{1: Uninhabitable} – Likely unsuitable for rehabilitation; abandoned, fire-damaged, boarded-up, or vacant. Requires demolition.
\item \textbf{2: Poor} – Requires substantial improvements, including major roof repairs, broken windows, bulging walls, or sagging foundations.
\item \textbf{3: Adequate} – Requires basic cosmetic repairs, with no more than two issues such as painting/siding, trim, porch, minor roof improvements, or fence repair.
\item \textbf{4: Good} – Structurally sound with good maintenance and no immediate repairs required. There may be no more than one minor issue, such as limited painting/siding replacement, minor porch repair/painting, or minor fence repair/painting.
\item \textbf{5: Excellent} – Recently rehabilitated or remodeled; no repairs needed. New paint and roof in very good condition.
\end{itemize}

Formatting instructions for the expected output are also indicated in the text prompt, enabling the model to evaluate different aspects of the building, including paint, window, structure, and maintenance, according to the criteria, and then provide an overall numerical rating. Figure~\ref{fig:rating-sample} shows five examples corresponding to ratings 1 to 5, respectively. This overall rating is directly compared with the MOS provided by human experts.

Within this framework, we employ two strategies to leverage Gemma 3: fine-tuning with expert supervision and knowledge distillation for efficiency.

%
First, to bring the model’s predictions closer to the mean opinion of human experts, we fine-tune a strong base model using a small set of human-labeled data. Given the substantial size of modern LLMs and the computational constraints of our local hardware environment, full fine-tuning is not feasible. To address this, we adopt a \textit{parameter-efficient fine-tuning} (PEFT)~\cite{Houlsby-ICML19} strategy, which selectively updates a small subset of the model’s parameters while keeping the majority fixed. Specifically, we apply \textit{quantized low-rank adaptation} (QLoRA)~\cite{Dettmers-NeurIPS23}, a technique that enables fine-tuning using low-precision weights and low-rank adapters. This approach significantly reduces memory consumption and training cost, while preserving model performance. 

%
Second, to enable efficient large-scale evaluation, we distill the capabilities of the largest Gemma 3 model (teacher) into smaller, more efficient models (students). For tasks requiring detailed text output, we train a smaller Gemma 3 to replicate the teacher’s reasoning and formatting. We transfer knowledge to lightweight vision models for rating-only tasks for faster inference. Distillation is performed using pseudo-labels generated by the teacher on unlabeled building images, removing the need for additional human annotation while preserving alignment with expert judgments.

\section{Experiments}

\subsection{Dataset and Metrics}

We collected 12,063 GSV images from six states: California, Florida, Georgia, Indiana, New York, and Texas. The images capture a wide range of building conditions and styles. We invited seven human raters to independently evaluate the building condition of 1,281 randomly selected images on a 1–5 scale. As a result, the labeled images received MOS ratings of 5 for 229 images, 4 for 582 images, 3 for 345 images, 2 for 105 images, and 1 for 20 images. The remaining 10,782 unlabeled images are used for knowledge distillation experiments. Though resolution varies due to cropping, all images are clear enough for reliable assessment, with a minimum resolution of 600$\times$300. 


\begin{table*}[htb]
\caption{Comparison of multimodal LLMs. We report SRCC and PLCC with respect to the MOS, along with inference speed (tokens per second), GPU memory usage (VRAM, in gigabytes), and the number of GPUs (nGPU) used to run each model. Inference speed is expressed in tokens per second, ensuring consistency across models regardless of the length of generated text. The best values are highlighted in bold.}
\label{tab:llm}
\centering
\setlength\tabcolsep{6pt}
{\footnotesize
\begin{tabular}{c|cc|ccc}
\toprule[1pt]
model & SRCC $\uparrow$ & PLCC $\uparrow$ & inference speed $\uparrow$ & VRAM $\downarrow$ & nGPU $\downarrow$ \\
\midrule
Mistral Small 3.2 24B & 0.63 & 0.67 & 33.86 & 23.75 & \bf{1} \\
LLaMA 4 Scout 17B$\times$16E & \bf{0.78} & \bf{0.79} & 37.28 & 64.58 & 2 \\
LLaVA 1.6 7B & 0.41 & 0.41 & 88.46 & 5.52 & \bf{1} \\
LLaVA 1.6 34B & 0.61 & 0.61 & 25.83 & 25.92 & \bf{1} \\
Qwen 2.5 VL 32B & 0.68 & 0.72 & 23.66 & 22.32 & \bf{1} \\
Qwen 2.5 VL 72B & 0.73 & 0.76 & 11.02 & 49.77 & 2 \\
Gemma 3 4B & 0.45 & 0.46 & \bf{97.37} & \bf{4.87} & \bf{1} \\ 
Gemma 3 27B & 0.77 & 0.77 & 26.64 & 18.48 & \bf{1} \\

\bottomrule[1pt]
\end{tabular}
}
\end{table*}

Since the label distribution is imbalanced (e.g., 1's and 2's are scarce), we emphasize correlation‑based evaluation, which is suitable for ordinal MOS labels and robust to skewed class frequencies. We employ the \textit{Spearman's rank correlation coefficient} (SRCC) to evaluate the alignment between model-predicted ratings and human assessments. This non-parametric metric measures the monotonic relationship between two ranked variables, capturing how well the predicted ratings preserve the relative ordering of the ground truth ratings. SRCC is especially well-suited for subjective tasks, where the precise numerical rating may vary across raters, but the relative ranking remains meaningful. In equation,
\begin{equation}
  \text{SRCC} = 1 - \frac{6 \sum_{i=1}^{N} (x_i-y_i)^2}{N(N^2 - 1)},
  \label{eqn:srcc}
\end{equation}
where $x_i$ and $y_i$ denote the model-predicted rating and MOS from human ratings for the $i$-th image, respectively, and $N$ denotes the number of images.
A higher SRCC value indicates stronger agreement in ranking between the model and human raters, with $\rho = 1$ indicating perfect rank correlation.

To assess the linear agreement between model predictions and human ratings, we also compute the \textit{Pearson's linear correlation coefficient} (PLCC). Unlike SRCC, which measures monotonic rank alignment, PLCC quantifies the strength of a linear relationship between predicted ratings and ground truth ratings. It is defined as
\begin{equation}
\text{PLCC} = \frac{\sum_{i=1}^{N}(x_i - \bar{x})(y_i - \bar{y})}{\sqrt{\sum_{i=1}^{N}(x_i - \bar{x})^2} \sqrt{\sum_{i=1}^{N}(y_i - \bar{y})^2}},
\label{eqn:plcc}
\end{equation}
where $x_i$ and $y_i$ denote the model-predicted rating and MOS for the $i$-th image respectively, and $\bar{x}$,  $\bar{y}$ are their corresponding means. $N$ denotes the number of images. A PLCC value of 1 indicates perfect linear correlation, 0 indicates no correlation, and -1 indicates perfect inverse correlation. PLCC is useful when assessing whether the predicted ratings follow the correct ordering and approximate the correct magnitude.


\subsection{Zero-shot Evaluation of Multimodal LLMs}
\label{subsec:llm}

We compare the zero-shot performance of multimodal LLMs on building condition evaluation and assess their inference costs. We select the latest open-source models from the LLaVA~\cite{Liu-NeurIPS23}, Mistral~\cite{Mistral}, LLaMA~\cite{Llama-arXiv23}, Qwen~\cite{Qwen-arXiv23}, and Gemma~\cite{Gemma-arXiv24} families. We evaluate two model sizes for LLaVA, Qwen, and Gemma to provide additional reference points. Experiments are run on a system equipped with four NVIDIA A40 GPUs (48 GB VRAM each). While most experiments are performed on a single GPU, models exceeding the memory capacity of one device in our system are run on two GPUs for inference. Note that the multi-GPU execution did not impact inference speed in our experiments. The test set comprises all 1,281 GSV images along with their MOS ratings. All models are given the same text prompt instructing them to evaluate paint, windows, structure, and maintenance, followed by an overall rating from 1 to 5 representing the building condition, which is compared with the MOS to calculate SRCC and PLCC. 

\begin{table*}[htb]
\caption{Comparison of individual human raters to the MOS, excluding each rater’s own ratings to ensure fairness. We report SRCC and PLCC for each rater, as well as the average across all human raters.}
\label{tab:human}
\centering
\setlength\tabcolsep{6pt}
{\footnotesize
\begin{tabular}{c|ccccccc|c}
\toprule[1pt]
rater & A & B & C & D & E & F & G & average\\
\midrule
SRCC $\uparrow$& 0.78 & 0.80 & 0.80 & 0.76 & 0.76 & 0.74 & 0.81 & 0.78 \\
PLCC $\uparrow$& 0.80 & 0.80 & 0.81 & 0.78 & 0.77 & 0.76 & 0.82 & 0.79 \\

\bottomrule[1pt]
\end{tabular}
}
\end{table*}

\begin{table*}[htb]
\caption{Comparison of different output formats. We report SRCC and PLCC, along with the average number of tokens in the response generated, the average response generation time (in seconds), and the prompt processing time per image (in seconds). The best values are highlighted in bold.}
\label{tab:gemma3-prompt}
\centering
\setlength{\tabcolsep}{4pt}
{\footnotesize
\begin{tabular}{lcccc}
\toprule[1pt]
output & details \& number & details \& word & single number & single word \\
\midrule
SRCC $\uparrow$ & 0.77 & 0.77 & 0.70 & \textbf{0.78} \\
PLCC $\uparrow$ & 0.77 & \textbf{0.78} & 0.66 & \textbf{0.78} \\
\# response tokens $\downarrow$ & 87.82 & 79.89 & \textbf{2.00} & \textbf{2.00} \\
response time $\downarrow$ & 3.29 & 3.00 & \textbf{0.08} & \textbf{0.08} \\
processing time $\downarrow$ & 1.06 & 1.04 & \textbf{0.96} & 1.13 \\
\bottomrule[1pt]
\end{tabular}
}
\end{table*}

As shown in Table~\ref{tab:llm}, LLaMA 4 Scout 17B×16E (`B' denotes the number of parameters in \textit{billions}, while `E' stands for \textit{experts} in a mixture of experts architecture) achieved the highest correlation with MOS, with an SRCC of 0.78 and a PLCC of 0.79, demonstrating strong agreement with human ratings, which has the same SRCC and PLCC as the average of human raters shown in Table \ref{tab:human}. However, because the model requires more than 64 GB of memory, it had to be run on two GPUs in our setup. In comparison, Gemma 3 27B followed closely in performance, achieving a value of 0.77 for SRCC and PLCC. Notably, it requires only 18 GB of VRAM, making it compatible with a single commercial-grade GPU and therefore more practical for deployment on local machines with constrained computational resources. In this comparison, Gemma 3 27B shows the closest alignment with human ratings for building condition evaluation on a single GPU, outperforming larger models such as Mistral Small 3.2 24B, LLaVA 1.6 34B, Qwen 2.5 VL 32B, and Qwen 2.5 VL 72B.

Performance consistently declines among LLaVA, Qwen, and Gemma as model size is reduced. For Qwen 2.5 VL, the drop in both SRCC and PLCC from 72B to 32B is less than 0.1, which is much smaller than the 0.2 drop observed from 34B to 7B for LLaVA 1.6 or the drop of more than 0.3 from 27B to 4B for Gemma 3. As a result, although Gemma 3 4B demonstrated the highest efficiency in inference speed and the lowest VRAM usage, its predicted ratings deviate substantially from human ratings, making it unsuitable for use.

Overall, Gemma 3 27B is the most practical out-of-the-box open-source choice for building condition evaluation on a single GPU, combining near-human MOS alignment with moderate speed and GPU memory requirements.

\begin{table*}[htb]
\caption{Comparison of different numbers of training images used to fine-tune the Gemma 3 27B model. We report SRCC and PLCC, along with the training time (in minutes). The best values are highlighted in bold.}
\label{tab:gemma3-ft}
\centering
\setlength\tabcolsep{6pt}
{\footnotesize
\begin{tabular}{c|ccccccccc}
\toprule[1pt]
\# training images & 0 & 100 & 200 & 300 & 400 & 500 & 600 & 700 & 800\\
\midrule
SRCC $\uparrow$& 0.78 & 0.77 & 0.78 & 0.80 & 0.81 & 0.82 & 0.82 & \bf{0.83} & \bf{0.83}\\
PLCC $\uparrow$& 0.79 & 0.78 & 0.79 & 0.81 & 0.81 & \bf{0.83} & 0.81 & 0.82 & 0.82\\
training time $\downarrow$& --- & \bf{4.52} & 8.20 & 12.68 & 16.57 & 21.48 & 25.38 & 30.52 & 34.68 \\
\bottomrule[1pt]
\end{tabular}
}
\end{table*}

\begin{table*}[htb]
\caption{Comparison of models fine-tuned on two training datasets labeled respectively by Gemma 3 27B and by human annotators. We report SRCC and PLCC for both datasets, along with the batch size used during training, inference speed (in images per second), and GPU memory usage at the inference stage (VRAM, in gigabytes). Inference speed is expressed in images per second because models other than Gemma 3 produce numeric outputs instead of text. The best values are highlighted in bold.}
\label{tab:gemma3-kd}
\centering
\setlength\tabcolsep{6pt}
{\footnotesize
\begin{tabular}{c|cc|cc|ccc}
\toprule[1pt]
& \multicolumn{2}{c|}{Gemma dataset} &\multicolumn{2}{c|}{human dataset}& batch & inference &\\
model & SRCC $\uparrow$ & PLCC $\uparrow$ & SRCC $\uparrow$ & PLCC $\uparrow$ & size & speed  $\uparrow$ & VRAM  $\downarrow$\\
\midrule
ResNet-50 & 0.68 & 0.66 & 0.52 & 0.53& 32 & 69.41 & 1.87 \\
MobileNetV3-L & 0.65 & 0.66 & 0.45 & 0.46 & 32 & \bf{69.91} & \bf{1.72} \\
EfficientNetV2-M & 0.73 & 0.74 & 0.60 & 0.61 & 16 & 31.01 & 2.09 \\
SwinV2-B & 0.73 & 0.74 & 0.61 & 0.63 & 16 & 32.61 & 2.36 \\
Gemma 3 4B & \bf{0.81} & \bf{0.80} & \bf{0.74} & \bf{0.73} & 1 & 3.05 & 9.10 \\
\bottomrule[1pt]
\end{tabular}
}
\end{table*}

\subsection{Optimization of Gemma 3}
\label{subsec:opt-gemma3}

Building on the findings in Section~\ref{subsec:llm}, we investigate flexible ways to leverage Gemma 3 for building condition evaluation. We first examine whether different output formats in the prompt affect the performance of the Gemma 3 27B model. As shown in Table~\ref{tab:gemma3-prompt}, we experiment with alternative formats for the overall rating, including using a word instead of a number, and restricting the model to output only a single number or word without additional text. This latter case is motivated by the fact that the number of tokens in the response directly impacts inference time. All experiments are evaluated on the same set of 1,281 images. The results in Table~\ref{tab:gemma3-prompt} indicate that when detailed descriptions are provided, the model's accuracy remains almost consistent regardless of whether the overall rating is expressed as a word or a number. When only the rating is needed, using a single word yields nearly the same accuracy but with much faster responses due to fewer generated tokens. However, accuracy drops significantly when restricted to outputting only a single number. We attribute this phenomenon to numbers with less semantic context than descriptive words, making it harder for the model to link to specific building conditions when used alone.

Next, we fine-tune Gemma 3 27B using MOS labels to improve its alignment with human ratings. Since the accuracy of the word-only response is comparable to that of detailed responses (see Table~\ref{tab:gemma3-prompt}), we adopt the word-only format for these experiments to simplify loss computation based on MOS, which is translated from numerical ratings into their corresponding descriptive words. From the dataset of 1,281 images, the first 800 images are allocated for training and the remaining 481 for testing. As shown in Table~\ref{tab:gemma3-ft}, we vary the number of training images by randomly sampling subsets from the 800 training images. The slightly higher PLCC for the pre-trained Gemma 3 27B model compared to earlier results is attributable to the change in the test set. During fine-tuning, the model is quantized to 4-bit precision to reduce memory usage. Following the Gemma 3 official guidelines, the LoRA configuration uses a scaling factor of 16 and a dropout rate of 0.05 to prevent overfitting. The rank of the low-rank matrix is set to 16, and the warm-up ratio is set to 0.03 to improve training stability. LoRA limits training to roughly 16\% of the model's parameters. The learning rate is set to $5\times10^{-5}$. The training batch size is 1, and only one epoch is run to avoid overfitting. The model is optimized with a next-token prediction loss, calculated as the cross-entropy between the predicted logits and the target labels.

Table~\ref{tab:gemma3-ft} shows that fine-tuning Gemma 3 27B on 500 labeled images yields performance that surpasses the MOS alignment of all individual human raters (see Table \ref{tab:human}). Increasing the number of training images beyond 500 does not yield definitive performance improvements, indicating that the model reaches a performance plateau. These results suggest that a fine-tuned Gemma 3 27B model trained on 500 images is sufficient to match or exceed human-level consistency, making it a practical replacement for manual rating in automated building condition evaluation.

\subsection{Knowledge Distillation using Fine-tuned Gemma 3}

We also experiment with response-based knowledge distillation using our Gemma 3 27B fine-tuned on 500 labeled images as a teacher model to fine-tune the Gemma 3 4B model, as well as several efficient vision models, including ResNet~\cite{He-CVPR16}, MobileNetV3~\cite{Howard-ICCV19}, EfficientNetV2~\cite{Tan-ICML21}, and Swin Transformer V2~\cite{Liu-CVPR22-ST}. For comparison, we prepare two training sets: one containing 10,782 building images labeled automatically by the fine-tuned Gemma 3 27B, and another containing 800 human-labeled images. Both are evaluated using the same 481 images. For the Gemma 3 4B model, we apply the same QLoRA fine-tuning procedure described in Section~\ref{subsec:opt-gemma3}. For the vision models, we optimize using \textit{mean squared error} (MSE) loss between the predicted ratings and the MOS, with a learning rate of $1\times10^{-4}$ during fine-tuning. The vision models are trained for up to 10 epochs on the Gemma 3–labeled dataset and up to 100 epochs on the human-labeled dataset. We identify the epoch that achieves the highest SRCC on the test set and report its performance.

As shown in Table~\ref{tab:gemma3-kd}, knowledge distillation effectively transfers the capabilities of the fine-tuned Gemma 3 27B model to smaller models. For all models, fine-tuning on the automatically labeled dataset leads to consistently higher performance than fine-tuning on the human-labeled dataset. The approach performs well across different architectures, including CNNs (ResNet-50, MobileNetV3-L, EfficientNetV2-M), transformers (SwinV2-B), and multimodal LLMs (Gemma 3 4B). This result indicates that the fine-tuned Gemma 3 27B model can serve as a dependable substitute for human annotators in creating a large-scale dataset for building condition evaluation, helping to overcome the scarcity of annotated data while significantly reducing labeling costs. Notably, EfficientNetV2-M and SwinV2-B achieve SRCC and PLCC values above 0.7, comparable to the base Gemma 3 27B model, while delivering more than $30\times$ faster inference. Fine-tuning Gemma 3 4B on the automatically labeled dataset enables it to achieve SRCC and PLCC values exceeding the base Gemma 3 27B model, with the added benefit of roughly $3\times$ faster inference. In this experiment, the fine-tuned LoRA adapters were not merged into the pre-trained model prior to inference. Instead, they were loaded separately at runtime, allowing us to keep only the compact LoRA files rather than full model checkpoints. This approach increased VRAM usage from less than 5 GB to over 9 GB, as both the base model and the adapter were held in memory during inference. Still, users can merge the adapters into the base model, reducing the VRAM consumption to a level close to the pre-trained model. 
Leveraging knowledge distillation from the fine-tuned Gemma 3 27B model enables scalable, fully automated evaluation with the flexibility to select models for different performance–efficiency trade-offs, making it practical to process datasets containing millions of images within reasonable timeframes.

\section{Extended Attribute Extraction}
\label{sec:eae}
In addition to predicting overall building condition, we examine whether multimodal LLMs can extract a broader set of housing attributes from street-view imagery. These attributes are intended to enrich the downstream visualization dashboard by describing not only the target house itself but also aspects of its visible surroundings. We formulate all attribute extraction tasks as multiple-choice question answering (QA) so that model outputs and human judgments are directly comparable. This setup allows us to evaluate three complementary properties: the stability of repeated LLM assessments, the inter-rater reliability of human experts and LLMs, and the alignment between LLM predictions and human expert assessments.

\begin{figure*}[t]
\centering
\includegraphics[width=\textwidth]{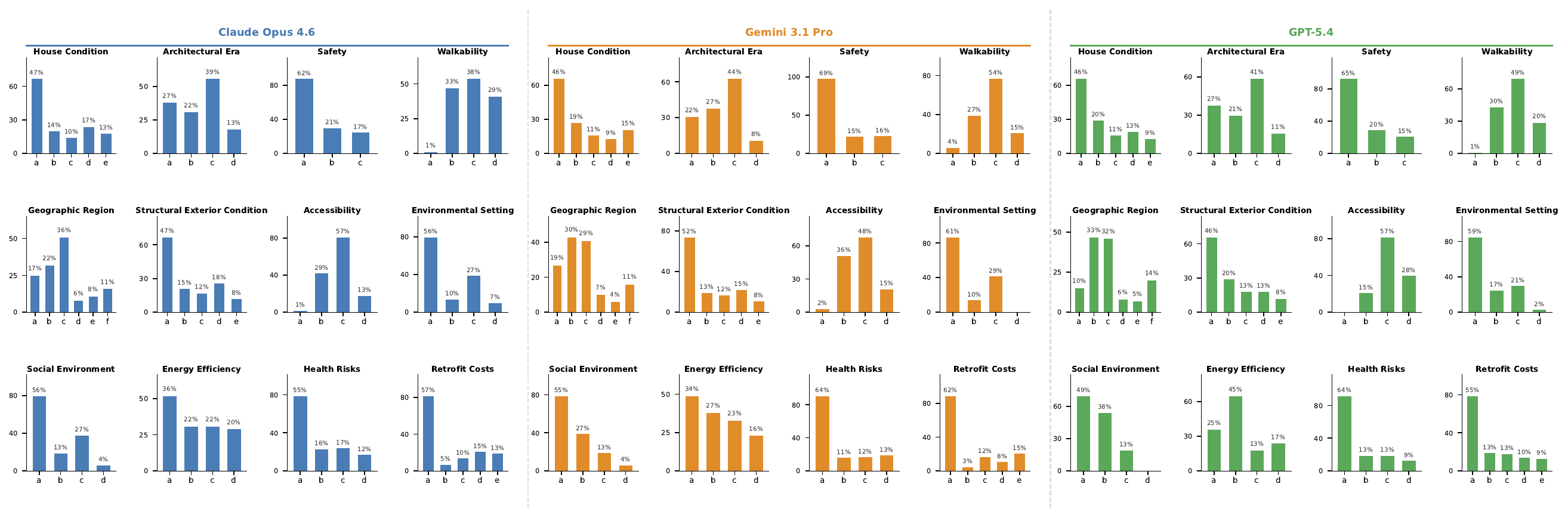}
\caption{Attribute-label distributions for GPT-5.4, Gemini-3.1-Pro, and Claude-Opus-4.6 after majority voting across five runs.}
\label{fig:dist-all-models}
\end{figure*}

\subsection{Attribute Extraction with Multimodal LLMs}
Unlike the experiments in Section~\ref{sec:approach}, which use cropped building views, this study uses zoomed-out street-view images that preserve the focal building together with its immediate context.
The goal is to determine whether multimodal LLMs can infer attributes that depend not only on the house itself, such as structural condition and architectural features, but also on visible neighborhood cues, including walkability, safety, accessibility, and social environment, as well as environmental context, including landscaping condition, environmental hazards, and geographic setting.
We collected 143 additional images spanning diverse geographic settings and housing conditions. Each image contains a single primary building while retaining enough surrounding context to support richer attribute assessment. For every image, we consider 12 attributes in total: the main housing-condition attribute and 11 additional attributes defined by architecture experts. The complete attribute definitions and prompting protocol are provided in Appendix~\ref{app:prompt-spec}.

We evaluate three proprietary multimodal LLMs: GPT-5.4, Gemini-3.1-Pro, and Claude-Opus-4.6. To ensure a fair comparison across models, we constrain each model to return exactly one categorical label for each attribute rather than an open-ended explanation. In other words, the task is cast as a multiple-choice QA problem. At inference time, the model receives the building image together with the 12 attribute questions and their candidate labels, and it must output one label per attribute. To reduce possible order effects, we randomly shuffle the order of the attributes for every run. Each model is queried five times per image, which enables us to evaluate both prediction quality and run-to-run consistency. Figure~\ref{fig:dist-all-models} shows the resulting attribute-label distributions for the three LLM judges.

\subsection{Human Expert Annotations}
To assess how closely LLM judgments align with expert opinion, we recruited six annotators with architecture expertise and asked them to complete the same multiple-choice task. Because this annotation process is substantially more time-intensive for humans than for models, we selected a representative subset of 25 images covering diverse geographic regions, neighborhood contexts, and levels of housing condition. The experts were shown the same zoomed-out images and the same attribute definitions used in model evaluation, and they assigned a single categorical label to each attribute for each image. This expert-labeled subset contains 300 image-attribute pairs and serves two purposes: it allows us to measure human inter-rater reliability and provides a reference for evaluating human-LLM alignment.

\subsection{Experimental Results}
We first evaluate whether each LLM is internally consistent across repeated trials with different random attribute orders. We define the \emph{stability score} as the average pairwise agreement across repeated runs. Specifically, for each image-attribute pair, we compare the model outputs across all $\binom{5}{2}=10$ run pairs and compute the fraction of run pairs that produce the same categorical label. We then average this fraction over all image-attribute pairs. A higher stability score, therefore, indicates that the model is less sensitive to prompt-order variation. As auxiliary dispersion summaries, for each image-attribute pair, we compute the standard deviation of the five repeated assessments on the ordered label scale, and then report the mean of these per-pair standard deviations over all image-attribute pairs. Lower values indicate more stable predictions. The results in Table~\ref{tab:judge_stability} reveal differences in model robustness to prompt-order variation. Claude-Opus-4.6 achieves the highest stability score and lowest variance, which aligns with its more concentrated distributions observed in Figure~\ref{fig:dist-all-models}. In contrast, GPT-5.4 exhibits the lowest stability and highest dispersion, and Gemini-3.1-Pro provides a balanced trade-off.

\begin{table}[htb]
\centering
\fontsize{8.5}{9.5}\selectfont
\caption{Stability of LLM assessments across 5 repeated trials with random attribute orders. A higher stability score indicates greater invariance to attribute ordering, while a lower mean standard deviation indicates less dispersion across runs. The best values are highlighted in bold.}
\label{tab:judge_stability}
\setlength{\tabcolsep}{2pt}
\renewcommand{\arraystretch}{0.98}
\begin{tabular}{@{}lccc@{}}
\toprule
model & \shortstack{stability score $\uparrow$} & \shortstack{avg std $\downarrow$} \\
\midrule
Gemini-3.1-Pro & 0.896 & 0.096 \\
Claude-Opus-4.6 & \textbf{0.941} & \textbf{0.054} \\
GPT-5.4 & 0.861 & 0.131 \\
\bottomrule
\end{tabular}
\renewcommand{\arraystretch}{1}
\setlength{\tabcolsep}{6pt}
\end{table}

Next, we compare the internal agreement of human experts and LLM judges using Krippendorff's $\alpha$ and the intraclass correlation coefficient, ICC(2,1). Krippendorff's $\alpha$ measures the extent to which raters assign consistent categorical labels, while ICC summarizes agreement when judgments are treated as ordered ratings. For both metrics, higher values indicate stronger inter-rater reliability. Table~\ref{tab:reliability_comparison} reports the agreement results for human experts, individual models, and the cross-model aggregate. Overall, all LLMs achieve substantially higher inter-rater reliability than human experts. Specifically, Krippendorff’s $\alpha$ and ICC values exceed 0.94 for each model and remain above 0.92 when aggregating across different models, indicating strong consistency both within and between LLMs. This suggests that LLM judgments are far more internally consistent than human annotations. However, higher agreement does not necessarily imply higher correctness. Instead, it reflects the more deterministic and less subjective nature of LLM outputs compared to human reasoning. Notably, Claude-Opus-4.6 achieves the highest agreement, reinforcing its strong stability observed in Table~\ref{tab:judge_stability}.

\begin{table}[t]
\centering
\fontsize{7.5}{8.5}\selectfont
\caption{Inter-rater reliability within human experts and LLMs (Gemini-3.1-Pro, Claude-Opus-4.6, GPT-5.4), measured by Krippendorff's $\alpha$ and ICC(2,1). Here, $n$ denotes the number of completed assessments, where each assessment is a full annotation over all 1,716 image-attribute pairs by one expert or one model run. The best values are highlighted in bold.}
\label{tab:reliability_comparison}
\setlength{\tabcolsep}{2pt}
\renewcommand{\arraystretch}{1.1}
\begin{tabular}{@{}lcc@{}}
\toprule
evaluator & \shortstack{Krippendorff's $\alpha$ $\uparrow$} & \shortstack{ICC(2,1) $\uparrow$} \\
\midrule
Human Experts ($n{=}6$) & 0.635 & 0.640 \\
Gemini-3.1-Pro ($n{=}5$) & 0.956 & 0.960 \\
Claude-Opus-4.6 ($n{=}5$) & \textbf{0.979} & \textbf{0.980} \\
GPT-5.4 ($n{=}5$) & 0.940 & 0.940 \\
Cross LLMs ($n{=}15$) & 0.920 & 0.918 \\
\bottomrule
\end{tabular}
\renewcommand{\arraystretch}{1}
\setlength{\tabcolsep}{6pt}
\end{table}

Finally, we evaluate how well LLM judgments align with human expert evaluations on the expert-annotated subset. For each image-attribute pair, human ratings are determined by majority voting across experts, and model predictions are determined by majority voting across repeated runs. We then compute Pearson correlation, Spearman rank correlation, {\em mean absolute error} (MAE), and {\em root mean square error} (RMSE) between the aggregated human and model judgments. Together, these metrics capture both rank-order consistency and absolute deviation from expert assessments.

As shown in Table~\ref{tab:human_llm_alignment}, all three models demonstrate strong alignment with human expert judgments, with correlation values above 0.78. Gemini-3.1-Pro achieves the best overall performance across both correlation and error metrics, indicating that it most closely matches human assessments despite not having the highest internal consistency. This suggests that higher stability or inter-model agreement does not necessarily translate to better alignment with human judgments. In contrast, Claude-Opus-4.6 shows slightly lower agreement with human annotations, indicating a possible bias toward more deterministic but less human-aligned predictions.

Overall, the results from Figure~\ref{fig:dist-all-models}, Table~\ref{tab:judge_stability}, Table~\ref{tab:reliability_comparison}, and Table~\ref{tab:human_llm_alignment} reveal a consistent pattern: models that produce more concentrated label distributions tend to achieve higher stability and inter-rater reliability, but not necessarily better alignment with human judgments. This suggests that different models occupy distinct points along a spectrum between consistency and human fidelity, providing flexibility for downstream applications depending on whether robustness or human alignment is prioritized.

\begin{table}[htb]
\centering
\fontsize{6.5}{8.5}\selectfont
\caption{Alignment between LLM assessments and human expert evaluations on the expert-annotated subset (25 images; 300 image-attribute pairs). Higher correlation and lower error indicate closer agreement with human assessments. The best values are highlighted in bold.}
\label{tab:human_llm_alignment}
\setlength{\tabcolsep}{2pt}
\renewcommand{\arraystretch}{0.95}
\begin{tabular}{@{}lcccc@{}}
\toprule
model & \shortstack{Pearson $r$ $\uparrow$} & \shortstack{Spearman $\rho$ $\uparrow$} & \shortstack{MAE $\downarrow$} & \shortstack{RMSE $\downarrow$} \\
\midrule
Gemini-3.1-Pro & \textbf{0.844} & \textbf{0.842} & \textbf{0.327} & \textbf{0.616} \\
Claude-Opus-4.6 & 0.803 & 0.800 & 0.423 & 0.705 \\
GPT-5.4 & 0.788 & 0.798 & 0.390 & 0.705 \\
\bottomrule
\end{tabular}
\renewcommand{\arraystretch}{1}
\setlength{\tabcolsep}{6pt}
\end{table}

\section{Visualization Dashboard}
\begin{figure*}
\centering
\includegraphics[width=\textwidth]{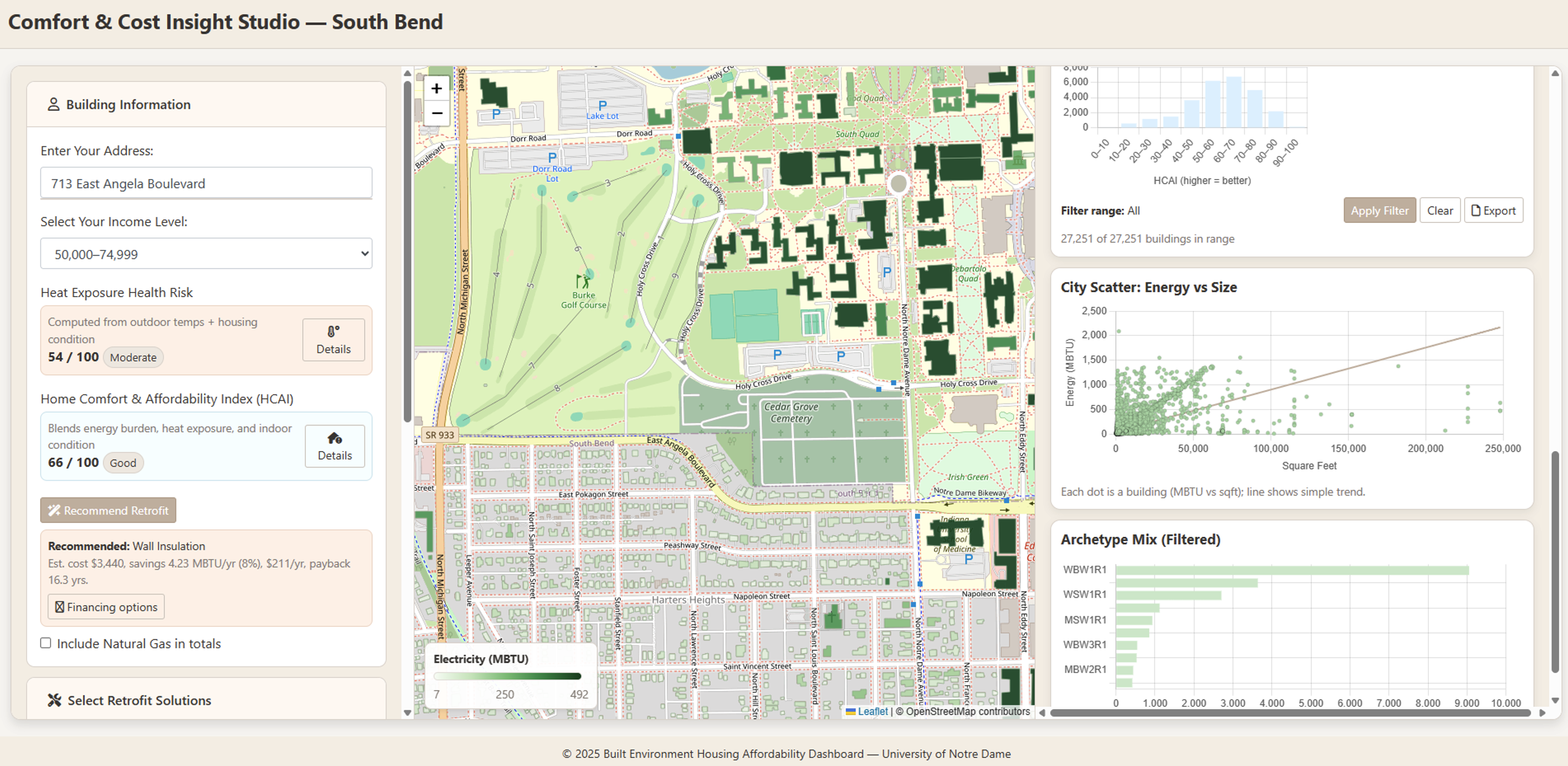}
\caption{Overview of the visualization dashboard, which integrates building-level property review, parcel-level map context, and city-scale comparison views in a unified interface.}
\label{fig:dashboard-overview}
\end{figure*}

While Sections~\ref{sec:approach}~to~\ref{sec:eae} focus on model development and evaluation, practical deployment also requires a user-facing interface to present these outputs in a form suitable for inspection and downstream decision-making. This motivates us to develop a visualization dashboard that connects the image-based building assessment with a broader building-analysis workflow. Figure~\ref{fig:dashboard-overview} shows a screenshot of the dashboard. At a high level, the dashboard integrates property information, map-based building lookup, LLM-assisted assessment results, and city-level comparison views in a single interface. The left panel supports building-level review and retrofit-oriented exploration, the central map provides parcel-level spatial context, and the right panel places the selected building within broader municipal patterns. In this paper, we include the dashboard to illustrate one possible way the proposed assessment pipeline can be presented to end users (e.g., homeowners), without attempting a full evaluation of the interface itself.

\begin{figure}[t]
\centering
\begin{subfigure}[t]{0.195\textwidth}
    \centering
    \includegraphics[width=\linewidth]{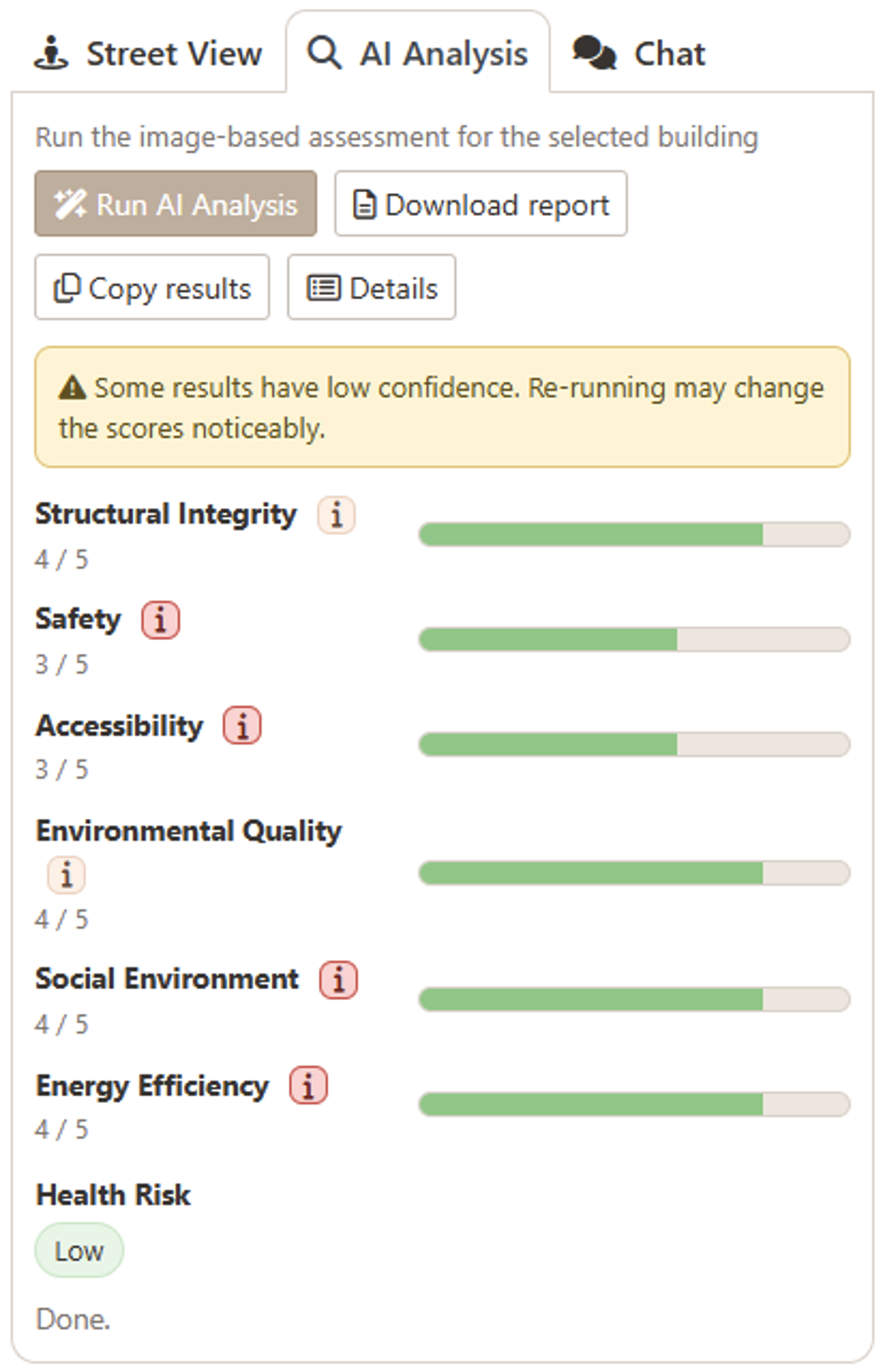}
    \caption{analysis tabs}
    \label{fig:overview1}
\end{subfigure}
\hfill
\begin{subfigure}[t]{0.265\textwidth}
    \centering
    \includegraphics[width=\linewidth]{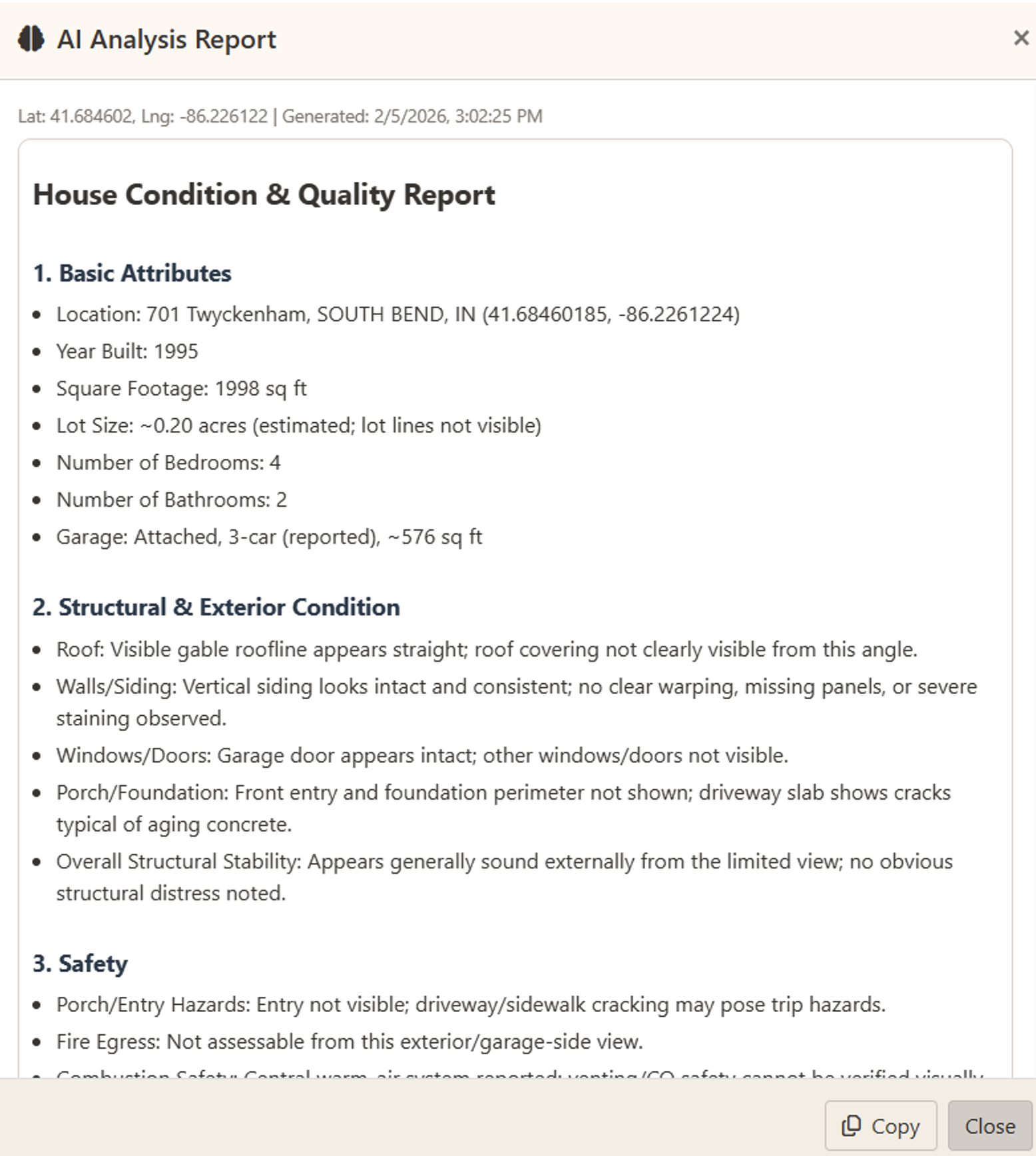}
    \caption{LLM report}
    \label{fig:overview2}
\end{subfigure}
\caption{Example dashboard views for LLM-based building assessment: the analysis tabs provide a compact summary of model outputs, while the report view presents detailed evaluation results for the selected property.}
\label{fig:detail-overview}
\end{figure}

Among the dashboard components, the AI analysis views shown in Figure~\ref{fig:overview1} and Figure~\ref{fig:overview2} are the ones most closely connected to this work. Building on the additional attribute extraction discussed in Section~\ref{sec:eae}, the dashboard combines street-view imagery with structured building attributes and neighborhood context, and then uses an LLM-based module to produce category-level assessments such as structural integrity, safety, accessibility, environmental quality, social environment, and energy efficiency. The interface is intended to make these outputs easier to inspect by presenting compact scores, confidence cues, and supporting details in a format that is more actionable than a standalone model prediction.

The report view in Figure~\ref{fig:overview2} complements the summary scores with a more explicit audit trail. Instead of returning only a single score, the system organizes basic property attributes, observed exterior conditions, and short textual rationales into a downloadable narrative report. This design is useful for practical review because it preserves the connection between the model output and the evidence available from street-view imagery and auxiliary records. At the same time, we keep the discussion here intentionally limited: components such as interactive chat, retrofit exploration, financing guidance, and broader city-level analytics are part of the dashboard environment, but a full evaluation of those modules is beyond the scope of this paper.

\section{Conclusions and Future Work}

We present a novel framework for automated building condition evaluation from GSV images across the United States using multimodal LLMs with minimal human annotation. We benchmark multiple leading open-source multimodal LLMs, identify the most effective model for aligning with expert ratings, and explore techniques, such as prompt refinement, targeted fine-tuning, and knowledge distillation, to enhance reliability and significantly improve efficiency, making million-scale dataset evaluation practical. The framework provides a flexible selection of methods, allowing users to balance performance and efficiency according to task-specific priorities. 
We further examine the capabilities of LLMs to assess an extensive list of built environment and housing attributes through a human–AI alignment study. Moreover, we develop a visualization dashboard that integrates LLM assessment outcomes for downstream analysis by homeowners.
While our work is the first to evaluate building quality at a large scale using LLMs, a promising future direction is to explore their application to broader aspects of buildings.

Still, our methods have a few limitations. 
First, the MOS ratings are derived from a specific group of skilled and novice raters, which may not fully capture broader subjective consensus or ensure complete fairness. While this may be sufficient for meeting the needs of a small target group, broader application scenarios require input from a larger and more diverse rater pool. 
Second, output-based knowledge distillation demands large quantities of raw images to generate pseudo-labeled image–rating pairs from real-world data. While feature-based knowledge distillation may offer greater efficiency, differences in network architectures introduce challenges in maintaining compatibility between the teacher and student feature representations. We consider exploring this as a future research direction. 
Third, given that LLM performance is highly dependent on prompt construction, future work will explore more sophisticated prompt engineering strategies beyond the plain-language criteria, query, and format descriptions used in this study. 
Finally, future work could examine potential biases in LLM outputs stemming from various image attributes, both low-level factors such as brightness and contrast, and high-level characteristics such as architectural style or building type, to better understand influences on model predictions.

In addition, future work will expand the visualization dashboard into a more thoroughly evaluated decision-support system. In particular, we plan to study how the AI analysis reports, additional attribute extraction, retrofit recommendation workflow, and city-scale visualization modules can be integrated and validated for different user groups, including residents, researchers, and local agencies.
\newpage
\bmhead{Acknowledgements}

This research was supported in part by the U.S.\ National Science Foundation through grants IIS-1955395, IIS-2101696, OAC-2104158, TI-2317971, IIS-2401144, and CNS-2430623, the University of Notre Dame's Just Transformations to Sustainability (JTS) and Data, AI, and Computing (DAC) Initiatives, Lucy Family Institute for Data \& Society Health Equity Data Lab, and Notre Dame-IBM Technology Ethics Lab.

\backmatter

\begin{appendices}

\section{Extended Attribute Extraction}
\label{app:prompt-spec}

To examine whether advanced closed-source multimodal LLMs can support broader housing analysis beyond the primary house-condition rating studied in the main paper, we prepared an additional prompt specification covering 12 attributes: one core housing-condition attribute and 11 additional contextual attributes. In practice, each model receives one street-view image together with the attribute definitions below and is asked to return exactly one label per attribute.

\subsection{Generic Evaluation Prompt}

The following template summarizes the prompt structure used for the three advanced closed-source models.

\noindent\textbf{Role.} You are an expert assessor of residential buildings from street-view imagery.

\noindent\textbf{Task.} Given a single exterior image of a house, evaluate the property using only visible evidence. Do not infer details that cannot be reasonably observed.

\noindent\textbf{Instructions.}
\begin{itemize}
    \item For each attribute below, choose exactly one label from the provided options.
    \item Base each judgment only on what is visible in the image.
    \item If evidence is limited, select the most conservative plausible label.
    \item Keep the output concise and structured.
\end{itemize}

\noindent\textbf{Required output format.}
\begin{itemize}
    \item House Condition: \textless label\textgreater
    \item Architectural Era: \textless label\textgreater
    \item Safety: \textless label\textgreater
    \item Walkability: \textless label\textgreater
    \item Geographic Region: \textless label\textgreater
    \item Structural \& Exterior Condition: \textless label\textgreater
    \item Accessibility: \textless label\textgreater
    \item Environmental Setting: \textless label\textgreater
    \item Social Environment: \textless label\textgreater
    \item Energy Efficiency: \textless label\textgreater
    \item Health Risks: \textless label\textgreater
    \item Retrofit \& Costs: \textless label\textgreater
\end{itemize}

\noindent Use the attribute definitions exactly as provided.

\subsection{Attribute Definitions}

\subsubsection{House Condition}
\begin{itemize}
    \item \textbf{Excellent:} Recently rehabilitated or remodeled; no repairs needed. New paint and roof in very good condition.
    \item \textbf{Good:} Structurally sound with good maintenance and no immediate repairs required. At most one minor issue is visible, such as limited painting or siding replacement, minor porch repair or painting, or minor fence repair or painting.
    \item \textbf{Adequate:} Requires basic cosmetic repairs, with no more than two issues such as painting or siding work, trim repair, porch repair, minor roof improvement, or fence repair.
    \item \textbf{Poor:} Requires substantial improvements, including major roof repairs, broken windows, bulging walls, or sagging foundations.
    \item \textbf{Uninhabitable:} Likely unsuitable for rehabilitation; abandoned, fire-damaged, boarded-up, or vacant, and likely requires demolition.
\end{itemize}

\subsubsection{Architectural Era (Estimated Year Built)}
\begin{itemize}
    \item \textbf{Pre-1950 (Historic):} Hand-laid brick or stone rather than veneer, true plaster, and wood windows or doors. Window openings tend to be smaller, with wood double-hung sash and divided panes. Taller-than-wide windows, visible chimneys, and crafted wood trim or porch elements are common.
    \item \textbf{1950--1980 (Mid-Century):} Long, low one-story ranches or split-level forms. Large single-pane picture windows with aluminum or steel frames, often in horizontal groupings. Minimal trim, low-pitch roofs, simple facades, and emphasized garages or carports are common.
    \item \textbf{1981--2010 (Modern Suburban):} Two-story massing with complex rooflines and multiple gables. Mixed-material facades such as brick or stone veneer combined with siding. Vinyl double-hung or casement windows, decorative grilles, arched shapes, and prominent attached garages are common.
    \item \textbf{2011--Present (Contemporary):} Simplified or boxy forms with cleaner lines, modern flat or low-slope roofs or refined suburban styles, fiber-cement or engineered siding, manufactured stone, high-contrast color palettes, and larger energy-efficient windows, often with dark frames.
\end{itemize}

\subsubsection{Safety}
\begin{itemize}
    \item \textbf{Secure:} Safe entryways, intact railings, and stable structural lines.
    \item \textbf{Compromised:} Minor hazards such as cracked steps, missing brick, or aging railings.
    \item \textbf{Hazardous:} Visible structural failures, sagging roofs, leaning columns, broken foundations, or blocked exits.
\end{itemize}

\subsubsection{Walkability}
\begin{itemize}
    \item \textbf{Pedestrian-Integrated:} Clear paved sidewalks on both sides of the street, with visible crosswalks or bike lanes.
    \item \textbf{Limited Connectivity:} Sidewalks are present but disconnected or only on one side of the street; the area remains primarily car-oriented.
    \item \textbf{Car-Dependent:} No sidewalks or footpaths are visible; properties transition directly into the street or a large parking lot.
    \item \textbf{Primitive / Rural:} Unpaved roads with no visible public infrastructure for pedestrians or transit.
\end{itemize}

\subsubsection{Geographic Region}
\begin{itemize}
    \item \textbf{Northeast}
    \item \textbf{Midwest}
    \item \textbf{Southeast / Gulf Coast}
    \item \textbf{Southwest / Desert (AZ, NM, NV)}
    \item \textbf{West Coast / Pacific Northwest}
    \item \textbf{Mountain / Northern Rockies (CO, UT, MT, ID, WY)}
\end{itemize}

\subsubsection{Structural \& Exterior Condition}
\begin{itemize}
    \item \textbf{Excellent:} Roof, siding, windows, and foundation all appear new or recently maintained, with no visible damage or wear.
    \item \textbf{Good:} Minor surface wear is visible on one element, such as slight paint fading or light weathering, with no structural concerns.
    \item \textbf{Fair:} Moderate deterioration is visible on one or two elements, such as an aging roof, worn siding, or hairline foundation cracks.
    \item \textbf{Poor:} Significant damage is visible across multiple elements, such as a sagging roof, broken windows, crumbling walls, or foundation issues.
    \item \textbf{Critical:} Severe structural-failure risk is visible, with near-collapse condition, extensive exterior damage, or major foundation failure.
\end{itemize}

\subsubsection{Accessibility}
\begin{itemize}
    \item \textbf{Fully Accessible:} Step-free entry, wide doorways, and a level approach are clearly visible.
    \item \textbf{Mostly Accessible:} One or two minor barriers are visible, such as a single step at the entry, but the property is otherwise manageable.
    \item \textbf{Limited Accessibility:} Several barriers are visible, such as multiple steps, a narrow entry, or no ramp or handrail.
    \item \textbf{Not Accessible:} Significant barriers are visible throughout, with no clear mobility accommodations.
\end{itemize}

\subsubsection{Environmental Setting}
\begin{itemize}
    \item \textbf{Well-Maintained:} The yard, landscaping, and surroundings appear neat and clean, with no visible environmental hazards.
    \item \textbf{Average:} Some yard maintenance is needed, but the surroundings are typical and show no obvious hazards.
    \item \textbf{Neglected:} The yard or surroundings appear overgrown, cluttered, or poorly maintained, with minor environmental concerns.
    \item \textbf{High Risk:} Visible environmental hazards are present, such as flooding indicators, industrial proximity, or a severely degraded lot or surrounding area.
\end{itemize}

\subsubsection{Social Environment}
\begin{itemize}
    \item \textbf{Strong:} Neighboring properties are well maintained, and clean streets or active landscaping suggest strong community investment.
    \item \textbf{Stable:} Nearby properties show mixed maintenance levels, but the neighborhood fabric appears generally intact.
    \item \textbf{Declining:} Several neglected or vacant neighboring properties are visible, with limited visible community investment.
    \item \textbf{Distressed:} Surrounding properties are predominantly vacant, boarded-up, or heavily deteriorated.
\end{itemize}

\subsubsection{Energy Efficiency}
\begin{itemize}
    \item \textbf{High Efficiency:} Newer construction or visible retrofit signs such as insulated siding, double- or triple-pane windows, or modern HVAC equipment are present.
    \item \textbf{Moderate Efficiency:} A mix of older and newer features is visible, with standard windows and typical systems for the era.
    \item \textbf{Low Efficiency:} Older single-pane windows, limited visible weatherization, or aging systems are apparent.
    \item \textbf{Very Low Efficiency:} The home appears pre-1950 with no visible upgrades and no evidence of weatherization or insulation improvements.
\end{itemize}

\subsubsection{Health Risks}
\begin{itemize}
    \item \textbf{Low Risk:} Surfaces are well maintained, with no visible mold, peeling paint, or moisture damage; the home may also be newer.
    \item \textbf{Moderate Risk:} Minor peeling paint or light moisture staining is visible, especially for older homes with mostly intact surfaces.
    \item \textbf{High Risk:} Visible peeling paint, potential mold or moisture staining, or significant deterioration is present on an older home.
    \item \textbf{Severe Risk:} Extensive deterioration is visible together with multiple health hazards, such as abandonment, fire damage, or partial collapse.
\end{itemize}

\subsubsection{Retrofit \& Costs}
\begin{itemize}
    \item \textbf{Move-in Ready:} No immediate retrofit is needed beyond cosmetic updates.
    \item \textbf{Light Retrofit:} Minor repairs are needed on one or two elements, implying low investment.
    \item \textbf{Moderate Retrofit:} Several components or systems visibly require replacement, implying moderate investment.
    \item \textbf{Heavy Retrofit:} Major structural, mechanical, or envelope work is clearly needed, implying significant investment.
    \item \textbf{Full Rehabilitation:} Near-complete renovation is required, corresponding to the highest investment level.
\end{itemize}

\end{appendices}

\bibliography{sn-bibliography}

@article{SG-ACMCS23,
    author = {Starzy{\'n}ska-Grze{\'s}, M. B. and Roussel, R. and Jacoby, S. and Asadipour, A.},
    title = {Computer Vision-based Analysis of Buildings and Built Environments: A Systematic Review of Current Approaches},
    year = {2023},
    volume = {55},
    number = {13s},
    pages = {284:1-284:25},
    journal = {ACM Computing Survey}
}

@inproceedings{He-CVPR16,
  title={Deep residual learning for image recognition},
  author={He, K. and Zhang, X. and Ren, S. and Sun, J.},
  booktitle={Proceedings of IEEE Conference on Computer Vision and Pattern Recognition},
  pages={770-778},
  year={2016}
}

@inproceedings{Liu-NeurIPS23,
  title={Visual instruction tuning},
  author={Liu, H. and Li, C. and Wu, Q. and Lee, Y. J.},
  booktitle={Proceedings of Advances in Neural Information Processing Systems},
   year={2023}
}

@article{Ghorbany-BE24,
    title = {Examining the role of passive design indicators in energy burden reduction: Insights from a machine learning and deep learning approach},
    journal = {Building and Environment},
    volume = {250},
    pages = {111126},
    year = {2024},
    author = {S. Ghorbany and M. Hu and S. Yao and C. Wang and Q. C. Nguyen and X. Yue and M. Alirezaei and T. Tasdizen and M. Sisk}
}

@article{Anacker-IJHP19,
    author = {Katrin B. Anacker},
    title = {Introduction: Housing affordability and affordable housing},
    journal = {International Journal of Housing Policy},
    volume = {19},
    number = {1},
    pages = {1-16},
    year = {2019}
}

@inproceedings{Hu-NSF25,
    title={{BUILT2AFFORD}: Machine-learning-driven passive retrofits for affordable housing},
    author={Hu, Ming and Ghorbany, Siavash and Yao, Siyuan and Wang, Chaoli and Sisk, Matthew},
    year={2025},
    booktitle = {Proceedings of Architectural Research Centers Consortium Annual Conference}
}

@article{Ghorbany-SR25,
  title={Data driven assessment of built environment impacts on urban health across {United States} cities},
  author={Ghorbany, Siavash and Hu, Ming and Yao, Siyuan and Sisk, Matthew and Wang, Chaoli and Zhang, Kai and Nguyen, Quynh Camthi},
  journal={Scientific Reports},
  volume={15},
  pages={19998},
  year={2025}
}

@inproceedings{Yao-ISVC24,
  title={Leveraging Zero-Shot Learning on Street-View Imagery for Built Environment Variable Analysis},
  author={Yao, Siyuan and Ghorbany, Siavash and Sisk, Matthew and Hu, Ming and Wang, Chaoli},
  booktitle={Proceedings of International Symposium on Visual Computing},
  pages={243-254},
  year={2024}
}

@inproceedings{Yao-ISVC25,
 author = "S. Yao and S. Ghorbany and M. Forstchen and A. Korotaszand and M. Sisk and M. Hu and C. Wang",
 title = "Leveraging Multimodal {LLMs} for Building Condition Assessment from Street-View Imagery",
 booktitle = "Proceedings of International Symposium on Visual Computing",
 year = "2025",
 pages = "219-231"
}

@article{Jiang-CACIE25,
  title={Large language model for post-earthquake structural damage assessment of buildings},
  author={Jiang, Yongqing and Wang, Jianze and Shen, Xinyi and Dai, Kaoshan},
  journal={Computer-Aided Civil and Infrastructure Engineering},
  year={2025}
}

@article{Hoang-CIN18,
  title={Image Processing-Based Recognition of Wall Defects Using Machine Learning Approaches and Steerable Filters},
  author={Hoang, Nhat-Duc},
  journal={Computational Intelligence and Neuroscience},
  volume={2018},
  pages={7913952},
  year={2018}
}

@article{Amrouni-PCF24,
  title={Next-generation building condition assessment: {BIM} and neural network integration},
  author={Amrouni Hosseini, Mani and Ravanshadnia, Mehdi and Rahimzadegan, Majid and Ramezani, Saeed},
  journal={Journal of Performance of Constructed Facilities},
  volume={38},
  number={6},
  pages={04024050},
  year={2024}
}

@article{Zou-ISPRS21,
  title={Detecting individual abandoned houses from {Google} street view: A hierarchical deep learning approach},
  author={Zou, Shengyuan and Wang, Le},
  journal={ISPRS Journal of Photogrammetry and Remote Sensing},
  volume={175},
  pages={298-310},
  year={2021}
}

@inproceedings{Liu-SIG23,
  title={A new approach to assessing perceived walkability: Combining street view imagery with multimodal contrastive learning model},
  author={Liu, Xinyi and Haworth, James and Wang, Meihui},
  booktitle={Proceedings of ACM SIGSPATIAL International Workshop on Spatial Big Data and AI for Industrial Applications},
  pages={16-21},
  year={2023}
}

@inproceedings{Wang-AGILE25,
  title={Can {CLIP} See Safe Streets? Comparing Human and {VLM} Perceptions of Walkability and Safety},
  author={Wang, Xinchen and Gilvear, Alesja and Li, Yijing and Ilyankou, Ilya},
  booktitle={Proceedings of AGILE Walking the X-min City Workshop},
  year={2025}
}

@inproceedings{Cheng-VTC24,
  title={Assessing urban safety: A digital twin approach using streetview and large language models},
  author={Cheng, Yuhan and Yin, Zhengcong and Li, Diya and Li, Zhuoying},
  booktitle={Proceedings of IEEE Vehicular Technology Conference},
  pages={1--5},
  year={2024}
}

@article{Li-arXiv24,
  title={{BuildingView}: Constructing Urban Building Exteriors Databases with Street View Imagery and Multimodal Large Language Mode},
  author={Li, Zongrong and Su, Yunlei and Wang, Hongrong and Zhao, Wufan},
  journal={arXiv preprint arXiv:2409.19527},
  year={2024}
}

@article{Malekzade-CEUS25,
  title={Urban attractiveness according to {ChatGPT}: Contrasting {AI} and human insights},
  author={Malekzadeh, Milad and Willberg, Elias and Torkko, Jussi and Toivonen, Tuuli},
  journal={Computers, Environment and Urban Systems},
  volume={117},
  pages={102243},
  year={2025}
}

@article{Liang-arXiv25,
  title={{OpenFACADES}: An Open Framework for Architectural Caption and Attribute Data Enrichment via Street View Imagery},
  author={Liang, Xiucheng and Xie, Jinheng and Zhao, Tianhong and Stouffs, Rudi and Biljecki, Filip},
  journal={arXiv preprint arXiv:2504.02866},
  year={2025}
}

@article{Gemma-arXiv24,
  title={Gemma: Open models based on {Gemini} research and technology},
  author={Mesnard, Thomas and Hardin, Cassidy and Dadashi, Robert and Bhupatiraju, Surya and Pathak, Shreya and Sifre, Laurent and Rivi{\`e}re, Morgane and Kale, Mihir Sanjay and Love, Juliette and Tafti, Pouya and others},
  journal={arXiv preprint arXiv:2403.08295},
  year={2024}
}

@article{Llama-arXiv23,
  title={{LLaMA}: Open and efficient foundation language models},
  author={Touvron, Hugo and Lavril, Thibaut and Izacard, Gautier and Martinet, Xavier and Lachaux, Marie-Anne and Lacroix, Timoth{\'e}e and Rozi{\`e}re, Baptiste and Goyal, Naman and Hambro, Eric and Azhar, Faisal and others},
  journal={arXiv preprint arXiv:2302.13971},
  year={2023}
}

@article{Qwen-arXiv23,
  title={Qwen technical report},
  author={Bai, Jinze and Bai, Shuai and Chu, Yunfei and Cui, Zeyu and Dang, Kai and Deng, Xiaodong and Fan, Yang and Ge, Wenbin and Han, Yu and Huang, Fei and others},
  journal={arXiv preprint arXiv:2309.16609},
  year={2023}
}

@misc{Mistral,
  author = {{Mistral AI Team}},
  title = {{Mistral Small 3}},
  year = {2025},
  url = {https://mistral.ai/news/mistral-small-3}
}

@inproceedings{Houlsby-ICML19,
  title={Parameter-efficient transfer learning for {NLP}},
  author={Houlsby, Neil and Giurgiu, Andrei and Jastrzebski, Stanislaw and Morrone, Bruna and De Laroussilhe, Quentin and Gesmundo, Andrea and Attariyan, Mona and Gelly, Sylvain},
  booktitle={Proceedings of IEEE International Conference on Machine Learning},
  pages={2790-2799},
  year={2019}
}

@inproceedings{Dettmers-NeurIPS23,
  title={{QLoRA}: Efficient finetuning of quantized {LLMs}},
  author={Dettmers, Tim and Pagnoni, Artidoro and Holtzman, Ari and Zettlemoyer, Luke},
  booktitle={Proceedings of Advances in Neural Information Processing Systems},
  pages={10088-10115},
  year={2023}
}

@inproceedings{Howard-ICCV19,
  title={Searching for {MobileNetV3}},
  author={Howard, Andrew and Sandler, Mark and Chu, Grace and Chen, Liang-Chieh and Chen, Bo and Tan, Mingxing and Wang, Weijun and Zhu, Yukun and Pang, Ruoming and Vasudevan, Vijay and others},
  booktitle={Proceedings of IEEE International Conference on Computer Vision},
  pages={1314-1324},
  year={2019}
}

@inproceedings{Tan-ICML21,
  title={{EfficientNetV2}: Smaller models and faster training},
  author={Tan, Mingxing and Le, Quoc},
  booktitle={Proceedings of IEEE International Conference on Machine Learning},
  pages={10096-10106},
  year={2021}
}

@inproceedings{Liu-CVPR22-ST,
  title={Swin {T}ransformer {V}2: Scaling up capacity and resolution},
  author={Liu, Ze and Hu, Han and Lin, Yutong and Yao, Zhuliang and Xie, Zhenda and Wei, Yixuan and Ning, Jia and Cao, Yue and Zhang, Zheng and Dong, Li and others},
  booktitle={Proceedings of IEEE Conference on Computer Vision and Pattern Recognition},
  pages={12009-12019},
  year={2022}
}

@article{gu2024survey,
  title={A survey on {LLM}-as-a-judge},
  author={Gu, Jiawei and Jiang, Xuhui and Shi, Zhichao and Tan, Hexiang and Zhai, Xuehao and Xu, Chengjin and Li, Wei and Shen, Yinghan and Ma, Shengjie and Liu, Honghao and others},
  journal={The Innovation},
  year={2026},
  note={In Press}
}

@inproceedings{zheng2023llmjudge,
author = {Zheng, Lianmin and Chiang, Wei-Lin and Sheng, Ying and Zhuang, Siyuan and Wu, Zhanghao and Zhuang, Yonghao and Lin, Zi and Li, Zhuohan and Li, Dacheng and Xing, Eric P. and others},
title = {Judging {LLM}-as-a-judge with {MT-Bench} and {Chatbot Arena}},
year = {2023},
booktitle = {Proceedings of Annual Conference on Neural Information Processing Systems},
pages = {46595-46623}
}

@inproceedings{shankar2024who,
author = {Shankar, Shreya and Zamfirescu-Pereira, J.D. and Hartmann, Bjoern and Parameswaran, Aditya and Arawjo, Ian},
title = {Who Validates the Validators? Aligning {LLM}-Assisted Evaluation of {LLM} Outputs with Human Preferences},
year = {2024},
booktitle = {Proceedings of ACM Symposium on User Interface Software and Technology},
pages = {131:1-131:14}
}

@article{chehbouni2025neither,
  title={Neither valid nor reliable? Investigating the use of {LLM}s as judges},
  author={Chehbouni, Khaoula and Haddou, Mohammed and Cheung, Jackie Chi Kit and Farnadi, Golnoosh},
  journal={arXiv preprint arXiv:2508.18076},
  year={2025}
}

@inproceedings{gebreegziabher-etal-2025-leveraging,
    title = "Leveraging Variation Theory in Counterfactual Data Augmentation for Optimized Active Learning",
    author = "Gebreegziabher, Simret A  and
      Ai, Kuangshi  and
      Zhang, Zheng  and
      Glassman, Elena  and
      Li, Toby Jia-Jun",
    booktitle = "Proceedings of Findings of the Association for Computational Linguistics",
    year = "2025",
    pages = "894-906"
}

\end{document}